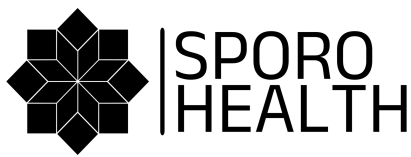

# Ambient AI Scribing Support: Comparing the Performance of Specialized AI Agentic Architecture to Leading Foundational Models

2nd investigation into comparing Sporo Health's algorithms to leading open- and closed-source LLMs.

**Authors:** Chanseo Lee BS,[1,2] Sonu Kumar MTech,[1] Kimon A. Vogt MS,[1] Sam Meraj MBBS[1]
1. Sporo Health, Boston, MA
2. Yale School of Medicine, New Haven, CT

## Abstract

*Background:* Integration of AI-powered tools into healthcare workflows necessitates a comparative understanding of Large Language Models (LLMs) for clinical documentation. This study evaluates Sporo Health's AI Scribe—a proprietary, multi-agentic model fine-tuned for medical scribing—against both open-source and closed-source LLMs, including GPT-4o, GPT-3.5, Gemma-9B, and Llama 3.2-3B.

Methods: We collected de-identified patient conversation transcripts from partner clinics using Sporo Health's proprietary transcription AI. Clinician-provided Subjective, Objective, Assessment, and Plan (SOAP) notes served as the ground truth. Each AI model was tasked with generating SOAP-formatted summaries from the transcripts using zero-shot prompting. The generated summaries were evaluated against the ground truth using recall, precision, and F1 score metrics. Clinical user satisfaction was assessed using the modified Physician Documentation Quality Instrument revision 9 (PDQI-9).

*Results:* Sporo's AI Scribe demonstrated superior performance, achieving the highest recall (73.3%), precision (78.6%), and F1 score (75.3%) among all models tested, with the lowest variance in performance. One-way ANOVA indicated statistically significant differences among the models (n = 10, $p < 0.05$). Post-hoc t-tests revealed that Sporo significantly outperformed GPT-3.5, Gemma-9B, and Llama 3.2-3B ($p < 0.05$). Sporo surpassed GPT-4o across all metrics by up to 10%, but the difference was not statistically significant ($p = 0.25$). PDQI-9 surveys showed that Sporo was significantly preferred. Evaluations of Sporo's outputs were notably more comprehensive and relevant, particularly in synthesizing medical information.

*Conclusion:* Sporo Health's AI Scribe outperforms both open-source and closed-source LLMs in accurately summarizing and reasoning through clinical data, as evidenced by objective metrics and clinician satisfaction. This underscores the high potential of Sporo's multi-agentic architecture and proprietary training methodologies to enhance clinical practice.

# Introduction

Following our initial case study, which compared the performance of Sporo AI Scribe against GPT-4o mini, we continue our exploration of AI-driven medical scribing tools by expanding our analysis to a broader range of models. In this second case study, we compare the performance of several open-source and closed-source Large Language Models (LLMs) in real-world clinical documentation tasks. Specifically, we assess GPT-4o,[1] GPT-3.5,[2] Gemma-9B,[3] and Llama 3.2-3B,[4] alongside Sporo's multi-agentic architecture, which has been specifically designed and fine-tuned for medical scribing.

As AI-powered tools become increasingly integrated into healthcare workflows, understanding the comparative strengths and weaknesses of these models is crucial. While the earlier case study highlighted the promise of AI scribing in improving clinician efficiency, accurate documentation remains a key challenge—especially when dealing with the complexity and nuance of medical language. A recent study by Sporo Health in the current state of AI adoption in clinical workflow reveals these crucial challenges in engineering and clinician satisfaction on the basis of clinical utility – namely in accuracy, comprehensiveness, and clinician-specific tailoring.[5]

Sporo Health's research-oriented approach to healthcare innovation produces key models and a new paradigm in AI applications – namely in its multi-agentic approach and pioneering training methodologies. In order to prove the effectiveness of our proprietary approach to healthcare AI, this study delves deeper into comparing Sporo's models against a broader range of available, closed-source or open-source models, evaluating how each model handles accuracy, contextual understanding, and the ability to capture all relevant details from patient encounters. With the growing adoption of AI scribing technologies and increasing traction for incorporating AI into clinical workflow, these comparisons offer insights into how different LLMs stack up in real-world medical environments, and the implications for future deployments in clinical settings. It is our hope to prove not only that Sporo Health is at the forefront of innovation, but also that AI is the next step to better healthcare.

## Methods

The following methods were also utilized in the first case study but expanded to larger samples. We collected a dataset of de-identified patient conversation transcripts from one of our partner clinics, generated using Sporo Health's proprietary speaker labeling and transcription AI. For each patient encounter, the piloting clinician provided Subjective, Objective, Assessment, and Plan (SOAP) notes through the scribe platform, which we designated as the ground truth. Sporo Health's AI agentic workflow, along with GPT-4o and GPT-3.5 hosted within Azure Playground, and natively hosted Gemma-9B and Llama 3.2-3B, were then tasked using zero-shot prompting with generating SOAP-formatted summaries from the same transcripts. These summaries were subsequently compared to the clinician's ground truth notes using various quantitative evaluation metrics.

**Clinical content recall (sensitivity)** was defined as the proportion of relevant clinical information from the clinician's ground truth summary that was accurately captured in the AI-generated summaries. To evaluate recall, salient clinical items were manually extracted from each conversation into an inventory. Recall was then calculated by dividing the number of correctly included items from the inventory by the total number of relevant items identified in the inventory.

**Clinical content precision (positive predictive value)** was defined as the proportion of information in the AI-generated summary that was both accurate and relevant when compared to the clinician's ground truth. Precision was calculated by dividing the number of correctly included items in the AI-generated summary by the total number of items in the AI summary, including any additional or incorrect items. This metric reflects the accuracy and relevance of the AI-generated content without introducing extraneous or inaccurate details.

The **F1 score** is used as a balanced metric to combine both clinical content precision and recall, providing a single measure of the AI-generated summaries' performance.[6] It represents the harmonic mean of precision and recall, ensuring that both the accuracy of relevant information captured (precision) and the completeness of that information (recall) are taken into account. The F1 score was calculated using the formula:

$$F1 = 2 \times \frac{\text{Precision} \times \text{Recall}}{\text{Precision} + \text{Recall}}$$

This metric is particularly useful for evaluating the overall effectiveness of the AI model when there is a need to balance precision and recall in the generated summaries.

In addition to the objective accuracy metrics, clinical user satisfaction was evaluated in conjunction with accuracy using the **modified Physician Documentation Quality Instrument revision 9 (PDQI-9)**. The original PDQI-9 employs a 5-point Likert scale across nine attributes to assess the quality of clinical notes. This was then modified by Tierney et al. into a ten-item inventory to better fit the metrics relevant to ambient AI documentation, and is widely used to evaluate AI-generated clinical notes.[7],[8] The attributes evaluated in the modified PDQI-9 are detailed in **Table 1.** Prior to evaluation, the conversation transcripts, the physician-generated summaries and the five associated AI-generated summaries were sanitized of any HIPAA-protected information.[9] A certified physician and a medical student first reviewed the physician-generated summary and conversation transcript, which served as the ground truth. They then reviewed the five AI-generated summaries—blinded to their source—and assessed the quality of the AI-generated notes using the PDQI-9. The evaluation particularly focused on both the quality of the notes and their alignment with the style, content, and utility of the physician-generated notes.

| PDQI-9 Attribute | Explanation |
| --- | --- |
| Accurate | The note does not present incorrect information. |
| Thorough | The information presented is comprehensive and lacks omissions. It contains all information that is relevant to the patient. |
| Useful | The information presented is relevant and provides valuable information for patient management. |
| Organized | The note is formatted in a way that is coherent and easy to comprehend. It helps the reader to understand the patient's story and the management of their clinical case. |
| Comprehensible | The note is straightforward, with no unclear or hard-to-understand sections. |
| Succinct | The note does not contain redundant information and presents relevant information in a concise, direct manner. |
| Synthesized | The note demonstrates the AI's comprehension of the patient's condition and its capability to formulate a care plan. |
| Internally Consistent | The facts presented within the note are consistent with each other and do not contradict the patient's story, each other, or known medical knowledge. |
| Free from Bias | The note is unbiased and includes only information that can be verified by the transcript, without being influenced by the patient's characteristics or the nature of the visit. |
| Free from Hallucinations | The information in the note aligns with the content of the transcript, without any factual inaccuracies or AI-generated hallucinations. |

**Table 1.** Items of the modified PDQI-9 for AI-generated summary evaluation.

## Results

Patient conversation transcripts (n = 10, avg. length = 3862 words) and their respective physician-generated notes (serving as the ground truth) were manually analyzed by an expert for salient clinical content items. The content included but was not limited to items relevant to the chief complaint (ex. "XX presents today for a follow-up appointment regarding their substance use disorder and to discuss treatment options"), social history (ex. social support, living situation, diet, occupation, exercise, relationships), patient emotions such as stress and anxiety, portions of the physical exam or discussions of lab results, and follow-up items such as scheduling an MRI. Each AI summary's averaged metrics across the summaries are shown in **Table 2,** with the best performing metric bolded.

|  | General Purpose | | | | Healthcare-Oriented |
| --- | --- | --- | --- | --- | --- |
|  | Closed Source | | Open Source | | Proprietary |
| Metric | GPT-4o | GPT-3.5 | Gemma-9B | Llama 3.2-3B | Sporo |
| Recall (%) | 67.1 | 61.6 | 33.4 | 60.3 | **73.3** |
| Precision (%) | 73.3 | 74.4 | 57.3 | 75.2 | **78.6** |
| F1 Score (%) | 68.9 | 66.0 | 39.8 | 65.6 | **75.3** |
| F1 σ (%) | 15.2 | 10.4 | 21.7 | 9.6 | **8.1** |

Table 2 Objective quantitative evaluation metrics for AI-generated summaries.

Sporo demonstrated superior accuracy metrics compared to other models, exhibiting the lowest performance variance as indicated by the standard deviation of the F1 score. A one-way ANOVA analysis revealed statistically significant differences in performance among the various models (n = 10, $p < 0.05$). Post-hoc two-tailed t-tests indicated that Sporo consistently outperformed GPT-3.5, Gemma-9B, and Llama 3.2-3B, with p-values below 0.05. Although Sporo surpassed GPT-4o across all evaluated metrics by a factor of **10%**, the observed differences did not reach statistical significance (p = 0.25), likely due to the limited sample size.

In addition to objective accuracy metrics, clinical utility and satisfaction was assessed using the modified PDQI-9 inventory and the averaged results from a clinician and medical student is shown in **Table 3.** Results showed that across the both evaluators, Sporo was significantly more preferred across accuracy metrics and overall medical utility. The results showed that there is a tradeoff between succinctness and thoroughness/accuracy – no models were limited in output size, but there is a preference for notes that can contain the most salient information in the least space.

| PQDI-9 | GPT-4o | | GPT-3.5 | | Gemma-9B | | Llama 3.2-3B | | Sporo | |
|---|---|---|---|---|---|---|---|---|---|---|
| Evaluator | A | B | A | B | A | B | A | B | A | B |
| Accurate | 4 | 4 | 3.5 | 3 | 2.5 | 3 | 3.5 | 3.5 | 3.5 | 4.5 |
| Thorough | 4 | 3.5 | 2.5 | 3 | 2 | 1.5 | 4 | 4.5 | 5 | 5 |
| Useful | 5 | 4 | 3.5 | 3.5 | 3 | 2 | 4 | 4.5 | 5 | 5 |
| Organized | 4.5 | 4 | 3 | 3.5 | 4 | 3 | 4.5 | 4 | 5 | 4 |
| Comprehensible | 4 | 4.5 | 4.5 | 5 | 3 | 3.5 | 4 | 5 | 4.5 | 5 |
| Succinct | 4.5 | 3.5 | 4 | 4 | 3 | 5 | 4 | 4 | 3.5 | 3.5 |
| Synthesized | 3.5 | 3.5 | 4 | 2 | 2 | 2.5 | 4 | 3.5 | 4.5 | 4 |
| Internally Consistent | 5 | 5 | 5 | 5 | 5 | 5 | 5 | 5 | 5 | 5 |
| Free from Bias | 4 | 3 | 4.5 | 3 | 5 | 3 | 4.5 | 3 | 5 | 3 |
| Free from Hallucinations | 3.5 | 3 | 4 | 3 | 4.5 | 2 | 3 | 3 | 3 | 3 |
| Total | 41 | 38 | 38.5 | 35 | 34 | 30.5 | 40.5 | 40 | 44 | 42 |
| Average | 39.5 | | 36.75 | | 32.25 | | 40.25 | | 43 | |

**Table 3.** Modified PDQI-9 evaluation of AI-generated summaries.

Lastly, the ability for each model to synthesize medical information accurately and produce outputs, specifically in the assessment, plan, and follow-up is crucial for the utility of the ambient AI assistants. In order to generate the quality notes with high accuracy and utility, AI should be able to reason through the details presented in each patient's clinical vignette and utilize it to 1. summarize information with all the salient pieces of information for a clinician's work and 2. aid in decision-making associated with diagnoses and management. Within the SOAP-structured notes that each model was asked to generate, Sporo in particular is trained to produce associated ICD-10 codes from presented medical information, summarize assessments, and suggest potential follow-up plans, which was noted by both evaluators and the clinicians utilizing the Sporo models in their clinic.

# Discussion

We present this case study to highlight Sporo's dedication to the highest quality of medical reasoning AI. Through this study, we were able to show using a comparative statistical model, applicable to real clinical practice, that Sporo can outperform both open-source (Gemma-9B, Llama 3.2-3B) and closed-source (GPT-4o, GPT-3.5) models in their ability to summarize and reason through real-world clinical data. This was proven through both objective accuracy metrics through manual clinical information inventories and clinician satisfaction surveys using the modified PDQI-9. This is one of many ways of proving that multi-agentic workflow and Sporo's proprietary methodologies in healthcare LLM training and deployment have the potential to transform clinical practice. It reflects Sporo Health's philosophy that AI should be more than a tool that a clinician must chaperone for full demonstration of utility – it should be an ambient assistant that ameliorates clinical outcomes and lessens burdens, not replace them.

Of particular interest to future studies are Meditron-7B and Mistral-7B, two open-source models trained for the purpose of medical applications. Recent evaluations of medical language models have highlighted notable differences in performance across various medical benchmarks. The Meditron-7B, a domain-specific model fine-tuned for healthcare tasks, has demonstrated strong results, achieving an average accuracy of 57.5% across key medical benchmarks, including MMLU-Medical (54.2%), PubMedQA (74.4%), MedMCQA (59.2%), MedQA (47.9%), and MedQA-4-Option (52.0%). In comparison, the Mistral-7B (including the Mistral-7B-instruct variant) performed relatively lower, with an average accuracy of 38.3% across the same tasks. Specifically, Mistral-7B showed strong results on MMLU-Medical (60.0%) but struggled significantly on domain-specific tasks such as PubMedQA (17.8%) and MedQA (32.4%). Meditron-7B showed a relatively stronger performance on domain-specific tasks, and also outperforms PMC-Llama-7B, a medically fine-tuned version of the popular LLaMA model, showcasing its advantage in specialized healthcare applications. These results underscore the effectiveness of Meditron-7B in handling complex medical queries and its superior performance in comparison to other general-purpose or fine-tuned models in the healthcare domain.[10]

We also hope to engage in future studies in patient chart review automation (pre-charting), conversion of language models to other languages, and studies with larger sample sizes and diverse ranges of specialties, which were currently limited due to the sheer quantity of work required for analyses like the ones presented in this case study.

# References

[1] OpenAI, Microsoft. GPT-4o (October 2024 version) [software]. San Francisco (CA): OpenAI; 2024 [cited 2024 Nov 6]. Available from: https://azure.microsoft.com/en-us/products/openai/

[2] OpenAI, Microsoft. GPT-3.5 [software]. San Francisco (CA): OpenAI; 2024 [cited 2024 Nov 6]. Available from: https://azure.microsoft.com/en-us/products/openai/

[3] Gemma Team. Gemma. Kaggle. 2024. Available from: https://www.kaggle.com/m/3301. doi:10.34740/KAGGLE/M/3301.

[4] Meta AI. (2024). Llama 3.2 3B: A multilingual large language model for dialogue and text generation tasks. [Computer software]. Meta. https://huggingface.co/meta-llama/Llama-3.2-3B

[5] Lee C, Vogt KA and Kumar S (2024) Prospects for AI clinical summarization to reduce the burden of patient chart review. Front. Digit. Health 6:1475092. doi: 10.3389/fdgth.2024.1475092

[6] Hicks SA, Strümke I, Thambawita V, Hammou M, Riegler MA, Halvorsen P, Parasa S. On evaluation metrics for medical applications of artificial intelligence. Sci Rep. 2022 Apr 8;12(1):5979. doi: 10.1038/s41598-022-09954-8. PMID: 35395867; PMCID: PMC8993826.

[7] Stetson PD, Bakken S, Wrenn JO, Siegler EL. Assessing Electronic Note Quality Using the Physician Documentation Quality Instrument (PDQI-9). Appl Clin Inform. 2012;3(2):164-174. doi: 10.4338/aci-2011-11-ra-0070. PMID: 22577483; PMCID: PMC3347480.

[8] Tierney AA, Gayre G, Hoberman B, et al. Ambient Artificial Intelligence Scribes to Alleviate the Burden of Clinical Documentation. NEJM Catalyst Innovations in Care Delivery. 2024;5(3). doi:https://doi.org/10.1056/cat.23.0404

[9] HIPAA PHI: Definition of PHI and List of 18 Identifiers. Human Research Protection Program - UC Berkeley. Published 2024. Accessed October 10, 2024. https://cphs.berkeley.edu/hipaa/hipaa18.html

[10] Chen Z, Hernández-Cano A, Romanou A, Bonnet A, Matoba K, Salvi F, Pagliardini M, Fan S, Köpf A, Mohtashami A, Sallinen A, Sakhaeirad A, Swamy V, Krawczuk I, Bayazit D, Marmet A, Montariol S, Hartley M-A, Jaggi M, Bosselut A. MEDITRON-70B: Scaling Medical Pretraining for Large Language Models. 2023. arXiv:2311.16079 [cs.CL].